\documentclass[conference]{IEEEtran}
\IEEEoverridecommandlockouts
\usepackage{cite}
\usepackage{subfigure}
\usepackage{amsmath,amssymb,amsfonts}
\usepackage{algorithmic}
\usepackage{graphicx}
\usepackage{textcomp}
\usepackage{url}

\usepackage{xcolor}
\def\BibTeX{{\rm B\kern-.05em{\sc i\kern-.025em b}\kern-.08em
    T\kern-.1667em\lower.7ex\hbox{E}\kern-.125emX}}
\begin{document}

\title{Regional Rainfall Prediction Using Support Vector Machine Classification of Large-Scale Precipitation Maps}

\author{\IEEEauthorblockN{ Eslam Hussein}
\IEEEauthorblockA{\textit{Department of Computer Science} \\
\textit{University of the Western Cape}\\
Cape Town, South Africa \\
eslamhuss34@gmail.com}
\and
\IEEEauthorblockN{ Mehrdad Ghaziasgar}
\IEEEauthorblockA{\textit{Department of Computer Science} \\
\textit{University of the Western Cape}\\
Cape Town, South Africa \\
mghaziasgar@uwc.ac.za}
\and
\IEEEauthorblockN{ Christopher Thron}
\IEEEauthorblockA{\textit{Department of Science and Mathematics} \\
\textit{University-Central Texas}\\
 Killeen, Texas, USA \\
thron@tamuct.edu}

}

\maketitle

\begin{abstract}
 
 Rainfall prediction helps planners anticipate potential social and economic impacts produced by too much or too little rain. This research investigates a class-based approach to rainfall prediction from 1-30 days in advance. The study made regional predictions based on sequences of daily rainfall maps of the continental US, with rainfall quantized at 3 levels: light or no rain; moderate; and heavy rain. Three regions were selected, corresponding to three squares from a $5\times5$ grid covering the map area. Rainfall predictions up to 30 days ahead for these three regions were based on a support vector machine (SVM) applied to consecutive sequences of prior daily rainfall map images. The results show that predictions for corner squares in the grid were less accurate than predictions obtained by a simple untrained classifier. However, SVM predictions for a central region outperformed the other two regions, as well as the untrained classifier. We conclude that there is some evidence that SVMs applied to large-scale precipitation maps can under some conditions give useful information for predicting regional rainfall, but care must be taken to avoid pitfalls.
 
\end{abstract}


\section{INTRODUCTION}
\subsection{Role of rainfall maps in water resource management}
Rainfall maps provide essential information about intensity, temporal, and spatial location which are essential in water resource management. Historical rainfall maps data can help different management sectors such as agriculture to make informed decisions about water supply management strategies to better utilize the occurrence of precipitation events \cite{harmel2003long}. Historical data can be most effectively utilized by developing prediction models such as machine learning to capture historical rainfall patterns.

In previous literature, prediction models based on rainfall maps may be grouped into two main categories. The first type involves applying deep learning to a sequence of images as an input to predict future frames. Usually, the images used for this type of prediction are separated by relatively small time intervals e.g 6-10 minutes \cite{xingjian2015convolutional, shi2018method, shi2017deep, tran2019multi, wang2017predrnn, singh2017deep, chen2020deep}.
The second type consists of single output regression or classification-based models. These models predict rainfall on an hourly \cite{kim2017deeprain, zhangtiny, yu2017comparison}, daily \cite{boonyuen2018daily}, \cite{boonyuen2019convolutional},  or monthly \cite{aswin2018deep} basis using prior rainfall maps. Regression-based models may use a single frame \cite{aswin2018deep} or batch of frames \cite{zhangtiny} as an input to give a numerical rainfall prediction. In contrast, classification based models categorize local rainfall into two or more discrete classes and predict the classes of future precipitation events based on a single frame or a batch of frames. 
\cite{boonyuen2018daily, chen2016short,boonyuen2019convolutional}.
Both regression and classification models can be used to predict entire images one pixel at a time \cite{yu2017comparison,mukhopadhyay2011novel }. 

\begin{figure*}
\centering
    \subfigure{\includegraphics[width=0.35\textwidth]{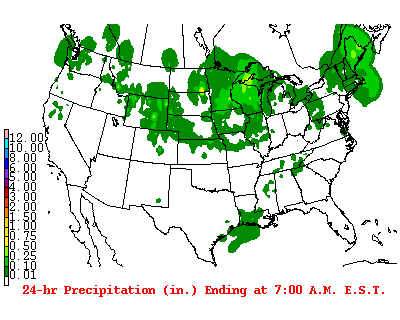}} 
    \subfigure{\includegraphics[width=0.30\textwidth]{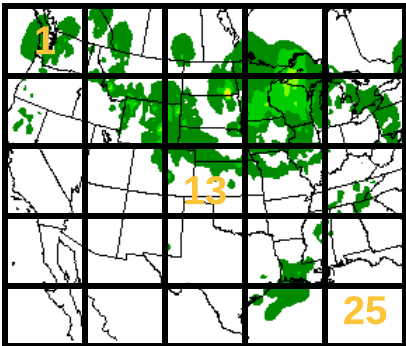}} 
    \subfigure{\includegraphics[width=0.32\textwidth]{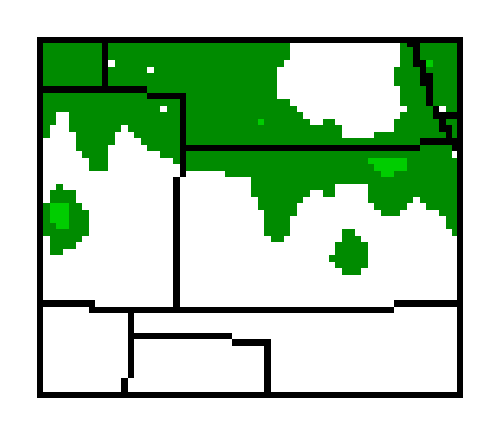}}  
    \caption{ Satellite precipitation images: (left) An image from the  NCEP dataset; (centre) {$5\time5$} grid overlay; (right) tile 13. }
    \label{24H}
\end{figure*}

For purposes of comparison, we describe the work of Boonyuen in \cite{boonyuen2018daily} and \cite{boonyuen2019convolutional}. In \cite{boonyuen2018daily} the authors used a single image to produce a binary classification (rain/no-rain) for three days ahead in Thailand. Using the inception-V3 based CNN model the authors had up to 54.84\% classification accuracy for three days ahead prediction. The study also concluded that including neighboring countries in the images increases the efficiency of the model compared to cropping the image to focus only on Thailand. 
In \cite{boonyuen2019convolutional} the authors developed an inception-V3 model to classify predicted rainfall into four categories (No-rain, light-rain, moderate-rain, heavy-rain). Both batches of satellite images and single images were used as input. The study demonstrated that using batches of images as input makes the model more robust at classifying upcoming rainfall. The trained model was able to predict one, two, three days ahead with an accuracy of 70.58\%. Having the same accuracy up to three days ahead is an issue of concern, as we suspect that the trained model has a bias towards the majority class (no-rain), making the accuracy to be constant. Measuring the efficiency of models using classification accuracy on imbalanced data is not ideal, because the results obtained may reflect the relative frequencies of the classes more than the actual effectiveness of the model. When imbalanced classes are involved, the f1-score can be a better measure of the method's effectiveness \cite{jeni2013facing}, as it takes the weighted average of precision and recall and is less influenced by class imbalance.

In The literature, various image sizes and sequence lengths are used in different prediction models. It appears that these parameters are usually chosen arbitrarily, or determined by trail and error. For example, the authors in \cite{zhangtiny} used up to 60 images to predict rainfall on an hourly base. the authors in \cite{shi2018method, shi2017convolutional, tian2019generative, singh2017deep,cao2019precipitation, chen2020deep}, used 4, 4, 5, 10, 10, 20, .20 images respectively. As to image size, the authors in \cite{zhangtiny} used several image sizes between $101 \times 101$ and $10 \times 10$, and compared the performance.

Even though most previous studies made use of deep learning related techniques to capture rainfall patterns on the historical images like Convolutional LSTM in \cite{xingjian2015convolutional}, in some cases the deep learning approach has significant drawbacks. In particular, as it often overfits when the training set is relatively small \cite{liu2015very, liu2017svm}. In addition, those models have many hyperparameters that need to be optimized. 

These difficulties can be avoided by using a support vector machine (SVM) approach instead of deep learning. SVM is a powerful machine learning technique where it is often used in classification and regression problems \cite{geron2017hands}. SVM is a classifier that generates a hyperplane to classify data instances \cite{liakos2018machine, manandhar2019data}, where optimal hyperplanes are determined by constructing the largest margin of separation between the different instances \cite{kim2003constructing}. In contrast to deep learning, SVMs are suited to be trained on small and medium-sized complex datasets \cite{geron2017hands, du2017prediction}. SVM also has the capability to perform structural risk minimization (SRM), which enables SVMs to avoid over-fitting by minimizing the bound on the generalization error \cite{chau2010hybrid}. 

Using SVMs can also avoid the need for extensive hyperparameter tuning. For example, the authors in \cite{hsu2003practical} considered the use of linear kernel (SVMs) in case of having the number of features exceeds the size of the dataset. Linear kernels use only a single parameter,  the regularization parameter $C$, that determines the trade-off between minimizing the training error and the model complexity \cite{duan2003evaluation}.

Several recent papers in the literature use SVMs on rainfall prediction for different classifications and regression problems. The authors in \cite{zainudin2016comparative} investigated the use of SVMs as well as other techniques to classify rainfall on a very small training set 10\% (2245), where the output was a binary classification rain/no-rain daily. Another binary classification problem (rain/no-rain) was studied in \cite{manandhar2019data}, which investigated the use of SVMs on weather stationary data to classify rainfall for the next five minutes. The data were highly imbalanced due to the rare occasion of rain, which made the researchers perform down-sampling on the dataset. As for regression problems, The authors in \cite{cramer2017extensive} used SVM to predict daily and the accumulated rainfall on 42 different cities from Europe and the US. The authors in \cite{yu2017comparison} investigated the use of SVMs with hourly radar-derived rainfall to predict precipitation during typhoons. Another study linked the observations from satellite imagery data to predict rainfall up to 6-hours \cite{chen2016short}. 

\subsection{Scope of this research}

\begin{figure*}
  \includegraphics[width=1\textwidth]{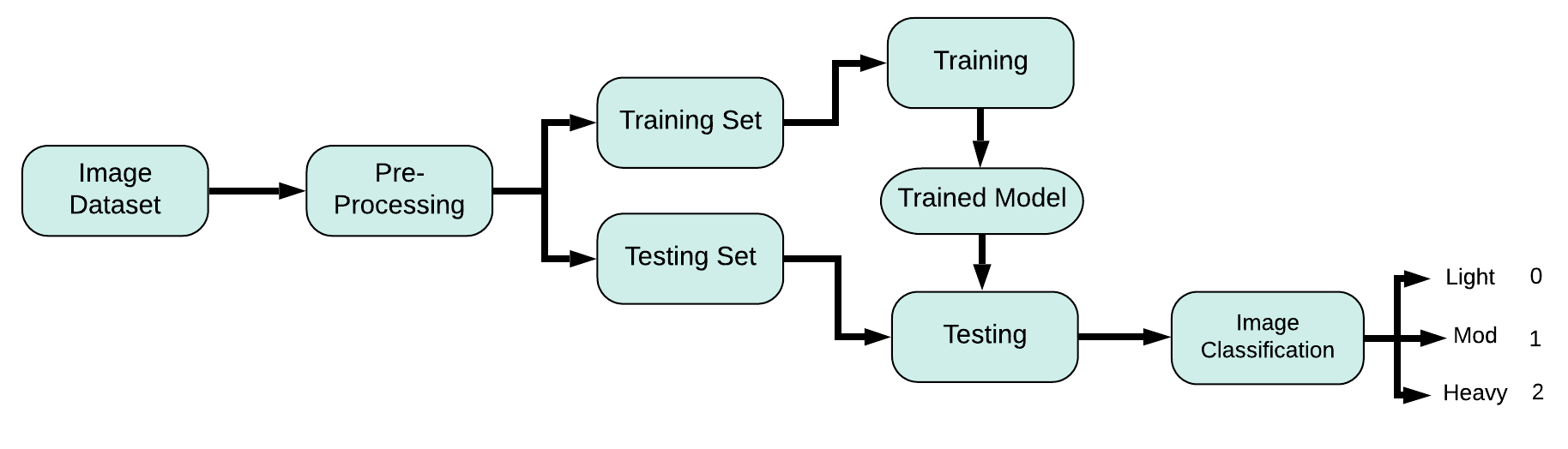} 
  \caption{Flow chart showing the implementation process.}
   \label{flowchart3}
\end{figure*}

\begin{figure}
    \centering
    \subfigure{\includegraphics[width=\columnwidth]{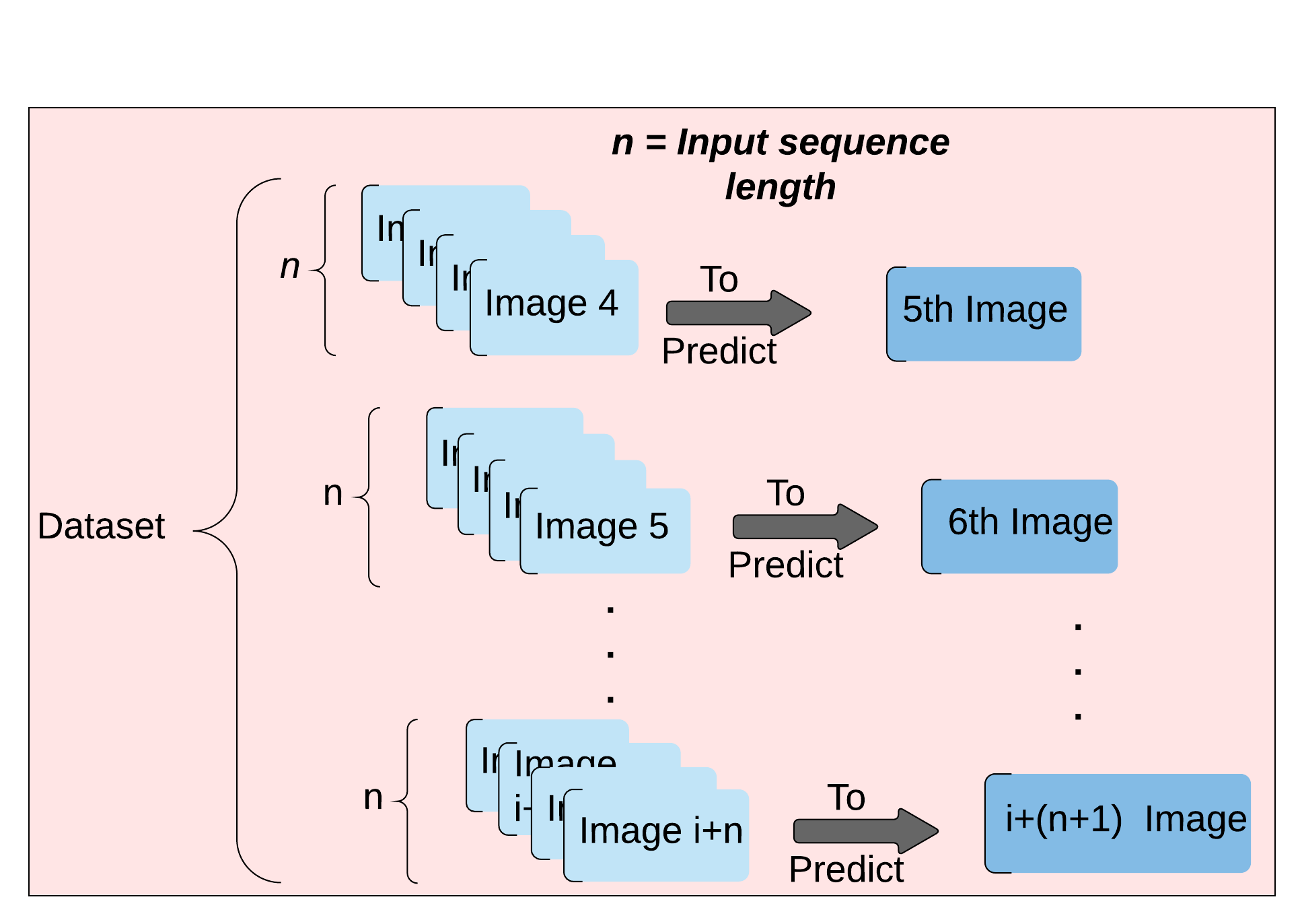}} 
    \caption{Overview of dataset preparation.}
    \label{dataPrep}
\end{figure}

This research aims to investigate precipitation forecasting on a  dataset from the National Center for Environmental Prediction (NCEP) using SVMs. Our investigation has three aspects: i) Determine  the  effect  of image sequence input length on  class  prediction  accuracy, ii) Assess the effect of image size on class prediction accuracy, and iii) Compare the accuracy of rainfall class predictions for three selected squares (tiles) from a $5 \times 5$ grid covering the map area, for up to 30 days ahead. 

This paper is organized as follows. Section II presents the methodology, including a discussion of the datasets and their preparation as well as the SVM specification and training. Section III presents the results in tabular and graphical form and provides analysis and discussion. Section IV summarizes our conclusions.

\section{Methodology}
Following the flow chart in Figure~\ref{flowchart3}, we start by discussing the data set, followed by the pre-processing of the images and the preparation of the data then explaining one of the models for this prediction. 

\subsection{Data Set}

\subsubsection{24-Hour-Precipitation-Forecasting}
The data used for this study are radar images taken daily at 7 a.m. Eastern Standard Time, from Jan 2012 to Oct 2019, with a total of 2,835 images. The data comes from the NCEP, with a size of 400$\times$320 pixels which represents the United States. Each image contains 16 different rainfall intensity level, Figure ~\ref{24H} (Left) shows a full image of the used dataset.
\newcommand\T{\rule{0pt}{4ex}}       
\newcommand\B{\rule[-2.3ex]{0pt}{0pt}} 

\begin{table*}[ht]
\centering
\caption{tile 1 (F1-score accuracy on the testing set). with $k$ Days Ahead (DA), with Different $n$ Input images and Sizes.}\label{tab1}
\begin{tabular}{cccccccccccc}
\hline
\textbf{Image} & \textbf{Input} & \multicolumn{10}{c}{\textbf{F1-score of Days Ahead (DA) (\%)}}\\
\cline{3-12}
\textbf{Scale} &     \textbf{Images} & 1DA        &    2DA        &     3DA & 4DA & 5DA & 6DA & 7DA  & 14DA & 30DA & Mean\\
\hline
{\textbf{Img(size \textit{a})}} 

&2& \textbf{60} &    31     &     27     &       30   &    26    &    23     &    \textbf{34}    &     22    & 22 &  30.55 \T\\
        
&4 & 52    &    37     &   \textbf{40}      &        26      &     24     &    32    &     22     &     22     & 22 & 30.77 \\

&6& 50 &     37      &     37     &         32     &     31   &   22     &     22    &     22     & 22 &30.66\\

&8 & 44     &    31     &    30    &       29      &  27     &    25     &    22    &    22     & 22  & 28\B\\

\hline

{\textbf{Img(size \textit{b})}}
&2 & 55 &    33   &   40    &   25    &   25  &   \textbf{ 33}     &    30    &    22   &    22 &31.66 \T\\

&4& 57 &   \textbf{43}   &   30     &   \textbf{35}    &  \textbf{ 37}  &  25     &    22     &   22    & 22 & \textbf{32.55} \\

&6 & 51   &    42     &   32    &       32    &    32     &    22     &   22    &   \textbf{29}     &    \textbf{23} &31.66\\

&8&   53   &     34     &    36     &       35     &    24    &    25     &    23   &    22     &  22 & 30.44\B\\
\hline

{\textbf{Mean}}
&  \T &   52.75   &     36     &   34     &      30.5     &   28.25    &   25.875     &    24.625   &    22.875     &  22.125  \B\\
\hline

\end{tabular}

\centering
\caption{tile 13 (F1-score accuracy on the testing set). with $k$ Days Ahead (DA), with Different $n$ Input images and Sizes.}\label{tab2}
\begin{tabular}{cccccccccccc}
\hline
\textbf{Image} & \textbf{Input} & \multicolumn{9}{c}{\textbf{F1-score of Days Ahead (DA) (\%)}}\\
\cline{3-12}
\textbf{Scale} &                 \textbf{Images} & 1DA        &         2DA            &           3DA &      4DA       & 5DA           & 6DA              & 7DA               &    14DA       & 30DA& Mean\\
\hline
{\textbf{Img(size \textit{a})}}
&2 &       51     &    49     &   \textbf{49 } &   42      &    42   &   42      &       43     &    38      &    44 & 44.44\T\\

                                            &4 &        58     &   \textbf{50}            &    45       &    41     &    44    &   \textbf{44}           &    45   &   46      &   \textbf{ 45} &\textbf{46.44}\\

                                            &6 &       51     &  50            &    42        &    39     &    44    &    44        &    40      &   \textbf{48}      &   41&44.33 \\

                                            &8 &        54   &   45            &   46        &   43     &   44     &   44           &   46    &   45      &    44 & 45.66  \B\\

\hline

{\textbf{Img(size \textit{b})}} 

&2         &       55     &   46    &    48               &  \textbf{ 46 }         &    41          &    40          &    \textbf{48}    &   46     &   37& 45.22\T\\

&4         &       55     &    47    &   43               &   45            &   43          &   41        &   43     &   43      &    40 & 44.44\\

&6        &       57     &    48     &   46               &   43            &         \textbf{ 49}        &    43         &   43       &    42    &    41& 45.77 \\

&8        &    \textbf{ 60  }   &    49         &   48           &    42               &    44    &    41           &    45     &   42   &   45  &46.22\B\\

\hline

{\textbf{Mean}}
&  \T &   55.125   &     48     &   45.875     &      42.625     &   43.875    &   42.375     &    44.125   &    43.75     &  42.125 \B\\
\hline

\end{tabular}

\centering
\caption{tile 25 (F1-score accuracy on the testing set). with $k$ Days Ahead (DA), with Different $n$ Input images and Sizes.}\label{tab3}
\begin{tabular}{cccccccccccc}
\hline
\textbf{Image} & \textbf{Input} & \multicolumn{9}{c}{\textbf{F1-score of Days Ahead (DA) (\%)}}\\
\cline{3-12}
\textbf{Scale} & \textbf{Images} & 1DA        &    2DA        &     3DA & 4DA & 5DA & 6DA & 7DA  & 14DA & 30DA & Mean\\
\hline
{{\textbf{Img(size \textit{a})}} }   
    
    &2 &        41     &    40     &    40    &    37     &    40     &   46     &    \textbf{49}    &   44     &   39 & 41\T\\
    
    &4 &      45     &   35    &    46     &    36     &  43     &   44    &    36     &   40    &   \textbf{49}&41.55 \\
    
    &6 &        46     &    48     &    39    &    46     &   39     &   41     &    39   &    42     &    49 &43.22\\
    
    &8 &      53     &    37     &    41     &    46     &    42     &    41    &    42     &   39     &    42& 42.55\B\\
\hline
    
{{\textbf{Img(size \textit{b})}}  }  
    &2 &       48     &   41     &    39              &    37              &    41            &   42             &   41     &   \textbf{45}     &    34&  40.88 \T\\
    
    &4 &       42    &   38     &    42               &   \textbf{48}               &   41     &         \textbf{ 47 }           &    41    &  42     &    42&  42.55 \\
    
    &6 &        \textbf{56}         &    \textbf{50 }    &   \textbf{ 49 }    &   44    &       \textbf{46}            &   39     &    41     &   41     &   41& \textbf{45.22} \\
    
    &8 &        49                   &    42              &   42               &   46     &         42     &    43    &   41     &   43     &   42 &  43.33\B\\
\hline

{{\textbf{Mean}}}
&  \T &   47.5   &    41.375    &   42.25    &      42.5     &  41.75    &   42.875     &    40.375   &   42     &  42.25 \B\\
\hline

\end{tabular}
\end{table*}

\begin{table*}[ht]
\centering
\caption{F1-score ranges for 8 different SVM inputs for different Days Ahead (DA), for three regions (\%)}\label{tab0}
\begin{tabular}{cccccccccc}
\hline
\textbf{ }   & \multicolumn{9}{c}{\textbf{F1-score range of Days Ahead (DA) (\%)}}\\
\cline{2-10}
     \textbf{Region}    & 1DA        &    2DA        &     3DA & 4DA & 5DA & 6DA & 7DA  & 14DA & 30DA\\
\hline

tile 1&   16 &    12     &     13     &       10  &    13    &    11     &    12    &     7    & 1  \T\\
        
tile 13  & 9    &    5     &   7      &       7      &    8     &    4    &    8     &     10     &8  \\

tile 25 &  15      &     15     &         10     &     12   &   7     &    8    &     6     & 6 & 15  \\

\hline

\end{tabular}
\end{table*}

\subsection{Pre-Processing}
The input data set went through a pre-processing transformation as shown in Figure~\ref{flowchart3}. The first transformation included size reduction, as we reduced the size of the images from $400 \times 320$ to $172 \times 123$ (image size \textit{a}), and $87 \times 61$ (image size \textit{b}). For both reduced data sets, we cropped the images to remove some of the irrelevant information that exists in our data set like the color bar on the left of Figure \ref{24H} (left map). In this study, we did not consider the full image size due to memory and time consideration.  After resizing we transformed the images into one channel by performing a grayscale transformation on the images to reduce the complexity of the model.

Figure \ref{dataPrep}, shows how we prepared the input dataset in a sliding window fashion to predict the next image in a sequence. Our use of sliding windows resembles the approach in several previous references \cite{qiu2017short, cramer2017extensive, chen2020deep, cao2019precipitation, klein2015dynamic}. The figure shows the case where $n = 4$, which stands for the size of the window (sequence length). In general, the number of feature (pixels) of one sample of the data set can be determined by $ n \times w \times h$, where $n$ is the size of the window, $w$ is the width of the image, and $h$ is the height of the image, which depends on the size of the images. 
For optimization, the study tested different size window sizes $n$, where $n$  $\in \{2,4,6,8\}$. We divided that data into 90\% training and 10\% testing for the whole experiment. Moreover, we divided the US map into a $5\times5$ grid squares (tiles), as We trained on specific grid tiles which are {1, 13, 25}.
\\
Initially, we divided the 16 classes on the color bar by Figure \ref{24H} left image equally into three classes {light, moderate, heavy}. For each image, tiles were classified according to the highest level observed within the tile: for example, if the tile had one or more pixels showing heavy rainfall, the entire tile was classified as heavy rain for that image.
However, this equal division produced highly unbalanced data, due to the rare occasion of very heavy rainfall. Consequently, our classification accuracy on the testing test was constant for predictions up to 30 days ahead, which is similar to what was observed in \cite{boonyuen2019convolutional} as discussed in the Introduction. To circumvent this problem, we made an unequal division between the classes by designating the lowest three classes as no/light rain, the next three classes as moderate rain, and all remaining classes as heavy rain. This improved the balance between the three classes: for the three tiles, we observed the following frequencies (no/light rain, moderate rain, heavy rain): (25\%, 49\%, 26\%) for tile 1, (38\%, 30\%, 32\%) for tile 13, and (36\%, 37\%, 27\%) for tile 25. 

\begin{figure*}
\centering
    \subfigure{\includegraphics[width=0.32\textwidth]{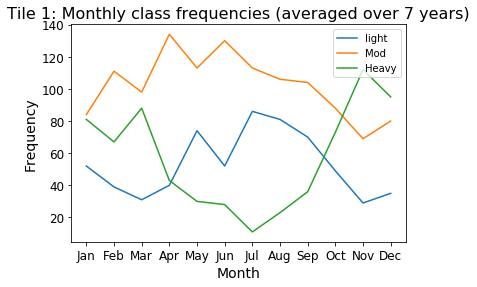}} 
    \subfigure{\includegraphics[width=0.32\textwidth]{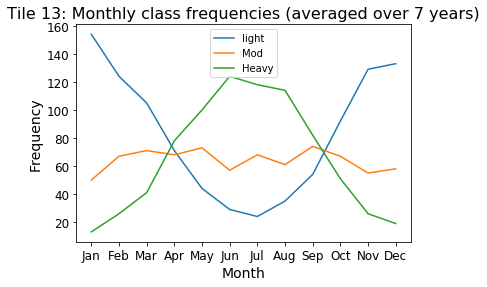}} 
    \subfigure{\includegraphics[width=0.32\textwidth]{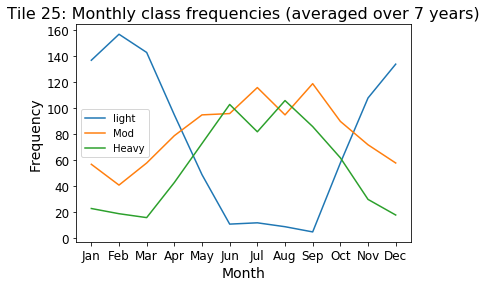}}  
    \caption{ Seasonality on a monthly basis for Tile 1 (\emph{left}), Tile 2 (\emph{middle}), and Tile 3 (\emph{right}).  }
    \label{fig:freq}
\end{figure*}

The three tiles show quite different seasonal behavior, as shown in Figure~\ref{fig:freq}. For Tile 13 (central), there is a clear distinction between the light rain and heavy rain class frequencies between summer months (May-Aug) and winter months (Oct-Mar). For Tile 25 (southeast), the light/no rain class shows strong seasonality, while the other two classes less so. For Tile 1 (northwest) the seasonality for all classes is less distinct. 

To summarise,  the input to our model is a set of full images, with different images sizes and windows (sequence length), while the prediction of the rainfall intensity happens on three specific tiles corresponding to three local regions within the U.S.

\subsection{Classification}
In this study, we used linear kernel SVMs, based on  \cite{hsu2003practical} which advised using a linear kernel when the number of features exceeds the size of the dataset.
These were trained on the training set, which comprised 90\% of the data. Training was accomplished using the \textit{sklearn.svm.SVC} class in scikit-learn (\url{www.scikit-learn.org}). Since we are working with a linear kernel, we had only one parameter to optimize which is the regularization parameter $C$. We tried values $C = 2^k$ for $k \in {-15,...6}$ with  10-fold cross-validation on the training set, and we chose the best \textit{C} value separately for each input configuration and each day ahead prediction, for each of the three tiles. (It was observed that in most cases the value of $C$ thus obtained was in the $2^{-13}$\textendash$2^{-10}$ range.) The models obtained were then applied to the testing set, and confusion matrices were obtained which were used to compute macro f1 scores. The macro f1-score was calculated as the average of the f1 scores of each class, where the per-class f1 score is computed as follows: 
\begin{equation}
  \text{f1-score} = \frac{2\times \text{Precision}\times \text{Recall}}{\text{Precision} + \text{Recall}}
 \label{eq:one}
\end{equation}

\section{Results and Discussion}

Tables \ref{tab1}, \ref{tab2},\ref{tab3} show the macro f1-score accuracy on the testing set for tile 1 (northwest), tile 13 (central) and  tile 25 (southeast) respectively. Each table gives results for different days-ahead predictions for 8 different SVM inputs (four different input sequences and two different image sizes). The bold numbers in each column represent the highest macro f1-score among the 8 SVM inputs for that specific days-ahead prediction for the given tile. 

In the following discussion, we will first compare the prediction performance for the different SVM inputs. Then we will compare the prediction performance for the three different geographical regions. 

\subsection{Comparison between different SVM inputs}
From Tables \ref{tab1}, \ref{tab2}, and \ref{tab3}  we do not find that any one input configuration is clearly better than the others. In Table \ref{tab1} for example, we find that 5 different input configurations attain the best accuracy for different days-ahead predictions. There is considerable variation within each column of the tables, as well as from column to column for each row. Table~\ref{tab0} shows the f1 score range (maximum $-$ minimum) among the 8 predictions for each days-ahead, for the three regions. From the table, it is clear that Tile 13, in general, has the lowest variability in the 1-6 day range, while the variability for Tile 1 reduces to almost 0 after 30 days. 

The observed variabilities may be attributed at least partially to the relatively small size of the testing set, which consists of about 300 images. For purposes of comparison, a sequence of 300 Bernoulli trials with success probability $p=0.5$ will have a standard deviation of $\pm 3$ percentage points. So a 95\% confidence interval of $\pm 2$ standard deviations will have a width of 12 percentage points. So in this Bernoulli trial scenario, the probability of getting 8 independent trials within a range of 12 is roughly $0.95^8 = 0.66$.    
 
\begin{figure}
    \centering
    \subfigure{\includegraphics[width=\columnwidth]{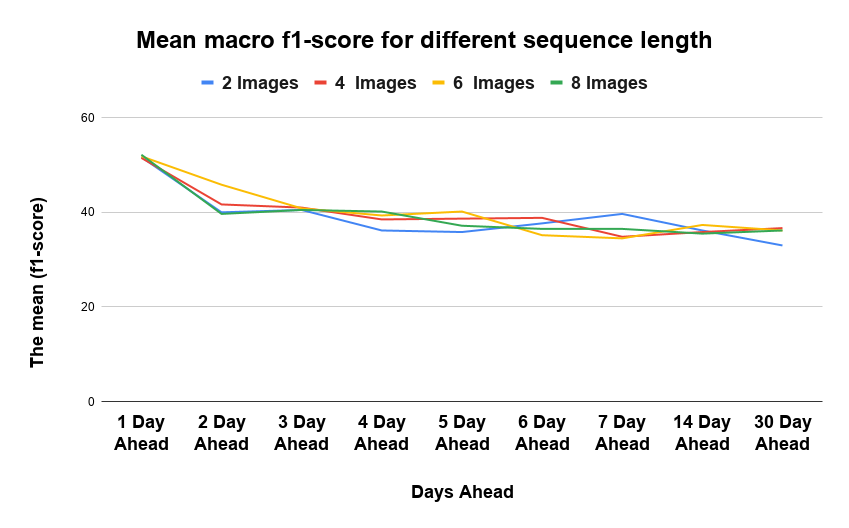}} 
    \caption{Mean macro f1-scores as a function of days ahead for different input image sequence length, averaged over all image sizes and tiles.}
    \label{G:len}
\end{figure}

\begin{figure}
    \centering
    \subfigure{\includegraphics[width=\columnwidth]{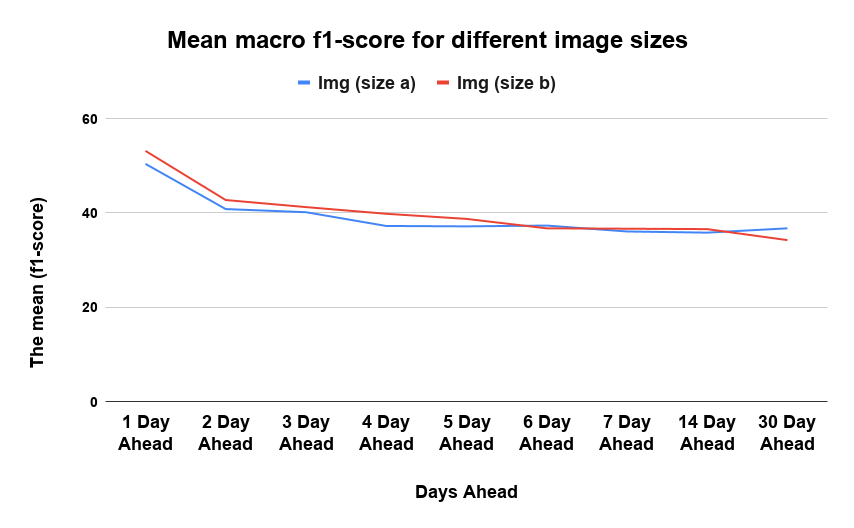}} 
    \caption{Mean macro f1 score as a function of days ahead for the two different image sizes, averaged over all input sequence lengths and tiles.}
    \label{G:size}
\end{figure}

 Figures~\ref{G:len} and~\ref{G:size} isolate the effect of input sequence length and image size, respectively. In Figure~\ref{G:len} the macro f1 scores for all predictions for all tiles for each days-ahead were averaged, and the results plotted as a function of days-ahead. The figure shows that no particular input sequence length is superior to the others. Figure~\ref{G:size} similarly averages macro f1 scores separately for each image size.  There appears to be a slight advantage of about 1 percentage point when using the larger image size ($172 \times 123$)  instead of the smaller size ($87 \times 61$).  Both figures show a clear decrease in accuracy as days-ahead increases. in contrast to the  constant classification accuracy found in \cite{boonyuen2019convolutional}.
 

\subsection{Comparison between regional predictions}

\begin{figure}
    \centering
    \subfigure{\includegraphics[width=\columnwidth]{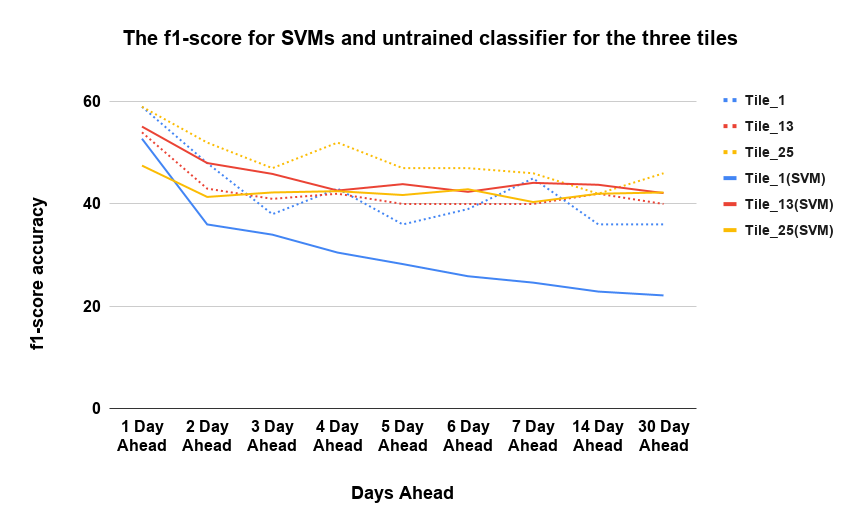}} 
    \caption{Per-tile average of f1-scores for 8 SVM inputs as a function of days ahead, compared to untrained benchmark predictor.}
    \label{G:compar}
\end{figure}

Figure~\ref{G:compar} shows the macro f1 scores averaged over 8 inputs (4 sequence lengths $\times$ 2 images sizes) for each day ahead, for each image separately. These f1 scores are compared to f1 scores obtained from an untrained predictor that simply uses the class of the final image in the sequence as the predicted class for $1,2,3, \ldots 30$ days ahead.  From the figure, it is clear that the SVM significantly underperforms the untrained benchmark predictor for Tiles 1 and 25. For Tile 13, the SVM outperforms the untrained predictor by about 1-3 percent. 

The SVM performance for Tile 1 is particularly poor, especially for longer-range predictions. Upon closer examination of the confusion matrices produced by the simulation, we found that for predictions longer than 4 days ahead, almost all of the SVM inputs were producing a classifier that always predicted the most frequent class (moderate rain) regardless of the SVM input. An elementary calculation based only on class frequencies (25\%, 49\%, 26\%) shows that a predictor which ignores inputs and always chooses the majority class will have a 
macro f1-score of 21.5\%, which is consistent with our observed result. 

The SVM performance for Tile 25 is also worse than the untrained predictor. The confusion matrices for Tile 25 shows that for larger-DA predictions, most predictors simply choose between light and moderate classes, and never predict the heavy class. This result is understandable  based on the seasonal behavior of Tile 25, shown in Figure~\ref{fig:freq} (\emph{right}). During the winter months (Nov.-Dec.), the light class is by far the most frequent. So if the input images show the tile as belonging to the light class, the chances are that the class will remain light for the next 30 days.  On the other hand, in the summer months (May-Sep), the light class is virtually nonexistent, and moderate and heavy are about equal.  Since moderate is a more frequent class than heavy (37\% versus 27\%), the classifier favors moderate over heavy.  Apparently, the untrained benchmark predictor more closely matches the seasonal pattern, which is why the untrained predictor performs better. We note that for Tile 25, a predictor that always chooses the majority class will have a macro f1 score equal to 18\%, so the SVM does represent a large improvement over a majority-class predictor.

Tile 13 showed the best SVM performance among the three tiles and was the only tile where the SVM outperformed the untrained benchmark classifier. We also noted from Table~\ref{tab0} that there was a smaller variation in macro f1 scores among the 8 different SVM inputs, for predictions between 1-5 days ahead. It is reasonable that the SVM's in Tile 13 are finding useful features and converging, while the SVM's for the other tiles are not locating truly useful features, so they are overfitting the training set which means that they no longer give consistent accuracy when applied to the testing set. It may be argued that Tile 13 has better input data than the other tiles because the input images contain information from all regions surrounding Tile 13 which is not the case for the other two tiles.

\section{Conclusion}
Our study points out some of the limitations and potential pitfalls in using SVMs with linear kernels for weather prediction up to 30 days in advance. We have shown that some classifiers used by previous researchers (e.g. \cite{boonyuen2018daily} and \cite{boonyuen2019convolutional}) which seem to show good performance may be largely due to unequal class divisions rather than the classifier itself. We have also shown that unequal classes may cause linear SVMs to converge on majority-class classifiers, or to completely neglect classes of low frequency. Among the three geographical regions predicted, only the central region had an SVM-based classifier that performed better than a simple untrained classifier that used the tile's class in the final image of the input sequence as the prediction. We conjecture that SVMs may be performing better on the central tile because the input sequence contains precipitation information for all surrounding tiles, which is not the case for tiles at the corners of the map (such as Tile 1 and Tile 25). 

In this study, we propose to divide the map of the US onto 25 tiles, as for the optimization we used different input configurations. The support vector machine was used to classify the image sequences as no/ light-rain: a) Our study shows that f1-score for tile 13 has a generally better f1-score than tile 1 and tile 25, which goes back to the position of the tile. b) Taking a bigger size scale as an input appears to provide slightly better performance than the smaller scale, but with higher time cost. c) The f1-score shows a decay while predicting days ahead, but the decay does not appear as prominent with imbalanced tiles e.g tile 1.

We have argued that the variability in the results obtained from day to day and from input method to input method was at least partially due to the insufficient amount of data for training and testing. Much of the variation in observed f1 scores is attributable to the small testing set size.  Recent research has shown that using data augmentation to augment the set of training images may improve the efficiency of the trained model\cite{tran2019multi}. With an augmented training set, a larger portion of the actual data may be used for testing. 

Another significant drawback of the SVM classifiers used was that they did not take seasonality into account. This is why a simple untrained classifier that took advantage of seasonality was able to substantially outperform the SVM classifiers on two out of the three regions examined. It is possible that SVMs that take seasonality into account my perform much better: this is a possible area for future research.

 In the current research we did not attempt to include additional engineered features, because the number of features used was already very large. For future investigation, a possible approach would be to use PCA to reduce the number of features, and then apply feature engineering. In addition to time stamp, Gaussian Mixture Models may be used to capture spatial means and variances, which could be used as global features.

\section*{Acknowledgment}

We acknowledge the use of the ilifu cloud computing facility- (\url{www.ilifu.ac.za})

\bibliographystyle{IEEEtran}
\bibliography{References}

\end{document}